\documentclass[]{spie}  %>>> use for US letter paper
\usepackage{amsmath,amsfonts,amssymb}
\usepackage{graphicx}
\usepackage[colorlinks=true, allcolors=blue]{hyperref}
\usepackage{bm}
\usepackage{wrapfig}

\title{Alternating Segmentation and Simulation for Contrast Adaptive Tissue Classification}

\author[a]{Dzung L. Pham}
\author[a]{Snehashis Roy}
\affil[a]{Center for Neuroscience and Regenerative Medicine, Bethesda, MD, USA}
\begin{document} 
\maketitle

\begin{abstract}
A key feature of magnetic resonance (MR) imaging is its ability to manipulate how the intrinsic
tissue parameters of the anatomy ultimately contribute to the contrast properties of the final,
acquired image.  This flexibility, however, can lead to substantial challenges for segmentation
algorithms, particularly supervised methods.  These methods require atlases or training data, which
are composed of MR image and labeled image pairs.  In most cases, the training data are obtained
with a fixed acquisition protocol, leading to suboptimal performance when an input data set that
requires segmentation has differing contrast properties.  This drawback is increasingly significant
with the recent movement towards multi-center research studies involving multiple scanners and
acquisition protocols.  In this work, we propose a new framework for supervised segmentation
approaches that is robust to contrast differences between the training MR image and the input image.
Our approach uses a generative simulation model within the segmentation process to compensate for
the contrast differences.  We allow the contrast of the MR image in the training data to vary by
simulating a new contrast from the corresponding label image.  The model parameters are optimized by
a cost function measuring the consistency between the input MR image and its simulation based on a
current estimate of the segmentation labels.  We provide a proof of concept of this approach by
combining a supervised classifier with a simple simulation model, and apply the resulting algorithm
to synthetic images and actual MR images.
\end{abstract}


\section{Description of Purpose}
\label{sec:intro}  % \label{} allows reference to this section

Magnetic resonance (MR) imaging plays a critical role in diagnosis, monitoring disease progression,
and response to therapies.  Automated segmentation algorithms for quantifiying brain structure in MR
images have shown increasing promise for routine application in both research studies and clinical
evaluations of neurodegenerative diseases.  The best performing algorithms currently rely on
training data or atlases to serve as exemplars of how MR images should be segmented.  Unfortunately,
such approaches perform suboptimally and inconsistently when faced with imaging data that possess
contrast properties that differ from the available exemplars.  This weakness serves as a substantial
disadvantage in longitudinal studies that are subject to scanner upgrades and continually evolving
imaging protocols.  Furthermore, in the age of ``big data,'' there is a huge demand to pool data
across multiple studies and sites for increasing statistical power.
% In a recent study examining the variability of
%volume variations in MS patients across scanners~\cite{Biberacher16}, MS patients were imaged on 3
%tesla (3T) MR scanners from different manufacturers.  Between scanner variation on a number of
%different brain and lesion volume measures were found to be significantly larger than within scanner
%variation.  
However, in a recent study scanning a patient at 7 different sites across North America, even with
careful matching of the MRI pulse sequence, variations across sites were
significant~[\citenum{ShinoharaIP}].

\begin{figure}[t]
\centering
\begin{minipage}{1\textwidth}
\tabcolsep 0pt
\begin{tabular}[ht]{cccccccc}
\includegraphics[height=2.6cm]{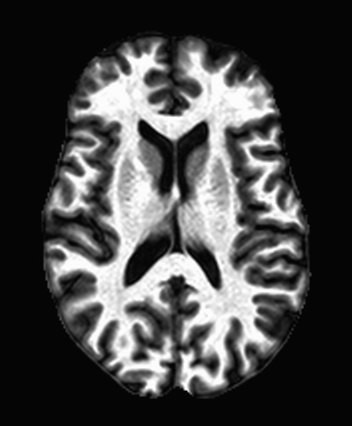}& 
\includegraphics[height=2.6cm]{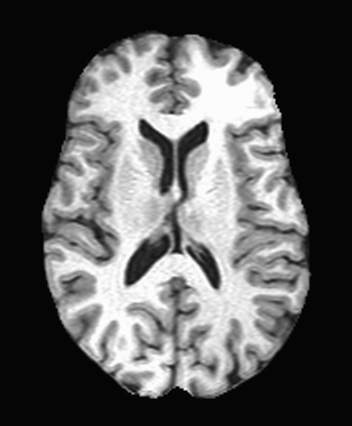}&
\includegraphics[height=2.6cm]{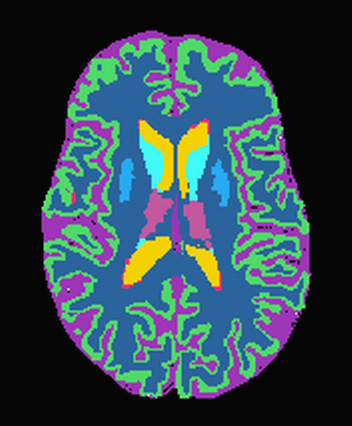}&
\includegraphics[height=2.6cm]{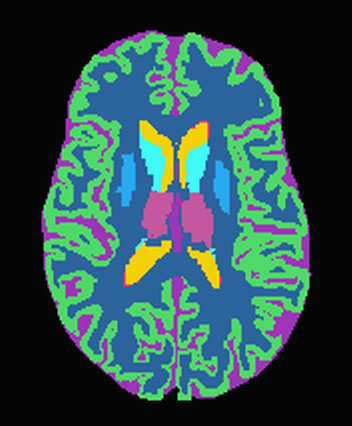}&
\includegraphics[height=2.6cm]{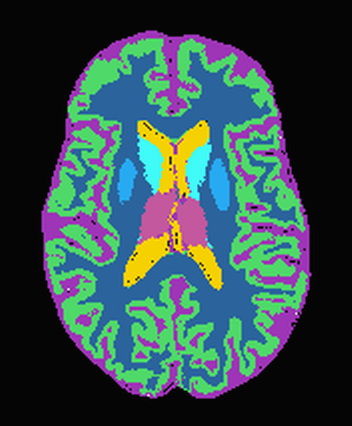} &
\includegraphics[height=2.6cm]{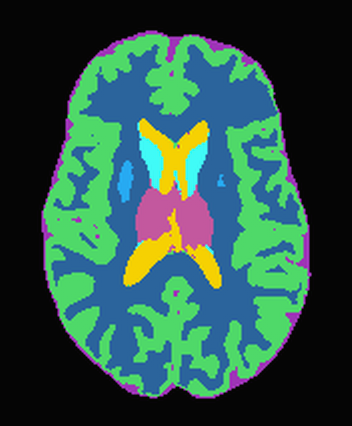}&
\includegraphics[height=2.6cm]{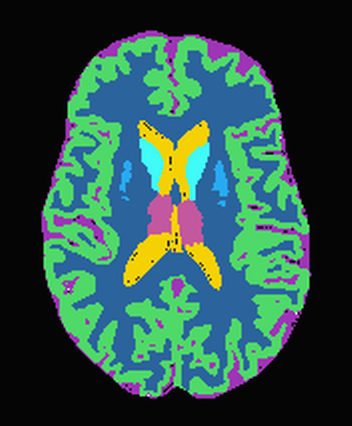}&
\includegraphics[height=2.6cm]{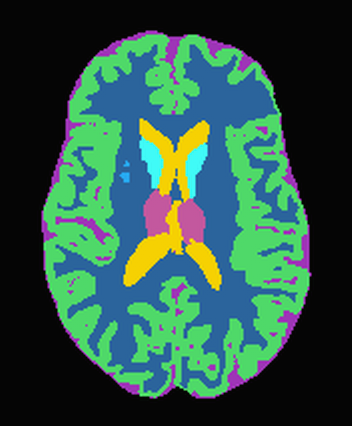}\\ 
\footnotesize T1-w A & \footnotesize T1-w B &
\footnotesize FreeSurfer A & \footnotesize FreeSurfer B& \footnotesize DL A & \footnotesize DL B &
\footnotesize LF A & \footnotesize LF B\\
\end{tabular}
\end{minipage}
\caption{\footnotesize Segmentation results given two different T1-w scans. DL stands for dictionary
  learning~\cite{Roy15c} and  LF stands for Label Fusion~\cite{Ledig15}.}
\vspace{-0.1in}
\label{fig:contrast_example}
\end{figure}

An example of the sensitivity of segmentation algorithms to MRI contrast is shown in
Fig.~\ref{fig:contrast_example}, where two T1-w images are acquired from the same subject.  Despite the
fact that the underlying anatomy should be the same, when three different segmentation algorithms
were applied (FreeSurfer~\cite{Fischl12}, S3DL~\cite{Roy15c}, and Label Fusion~\cite{Ledig15}), each
yielded substantially different results depending on the contrast.  In all cases, the second
T1-weighted image yielded a thicker cerebral cortex with smaller sulcal gaps.  Note also that the
different algorithms exhibit different sensitivities to the contrast differences, with the S3DL
approach showing the most dramatic differences.  As a patch-based approach, it is more dependent on
the intensity variations of its training data, leading to greater sensitivity to contrast
differences.

Multiple approaches have been previously proposed to perform contrast or intensity normalization.
The most common approach to compensating for site differences is to employ a site covariate within
the statistical modeling~[\citenum{Jones13,Chua15}].  However, this has not been well validated,
and there is limited evidence to suggest that imaging variations are well captured by such an
approach.
%, as the effects of acquisition differences are likely to be nonlinear and varying by spatial or
%anatomical region.  
Intensity normalization techniques have also been
proposed~[\citenum{nyul2000new,Shinohara14}] that attempt to align the histograms of images
using linear or piecewise linear transformations.  Because the transformations affect the global
histogram, local contrast differences that are region specific are not addressed by this approach.
Furthermore, there is an inherent assumption that the global histogram is similar in the two images
being matched.  In the case where one image possesses pathology and the other does not, this
assumption is violated.

%Phantom calibration has also been
%proposed~\citenum{Gunter09,Droby15}, but there inevitably exist significant differences between the
%phantom and actual human anatomy.  Such calibrations also cannot be performed retrospectively on
%legacy data, and can be impractical to perform prospectively with sufficient frequency in large-scale
%studies.  Transfer learning has been shown to improve the robustness of machine learning algorithms
%to contrast differences~\citenum{Opbroek15}.  Such techniques are complementary to the methods proposed
%here.

%Tissue specific approaches have been proposed that perform regional intensity transformations based
%on segmentation masks~\citenum{Robitaille12}.

Image synthesis is a class of image processing techniques where images from a subject are used in
combination with training data to create a new image with desirable intensity
properties~[\citenum{Roy13b,Jog15,Jog17}].  A matching process is performed between the input images
and the training data images with the same contrast, and an output image with the desired contrast
properties is computed.  The advantage of image synthesis for contrast normalization over histogram
matching approaches is that the transformation can be based on local properties of the images, and
is non-parametric.  Because of this greater flexibility, synthesis has been shown to outperform
global histogram normalization methods for contrast normalization~[\citenum{Jog15}].

\begin{wrapfigure}{r}{0.3\textwidth}
\vspace{-0.2in}
\begin{center}
\includegraphics[width=5cm]{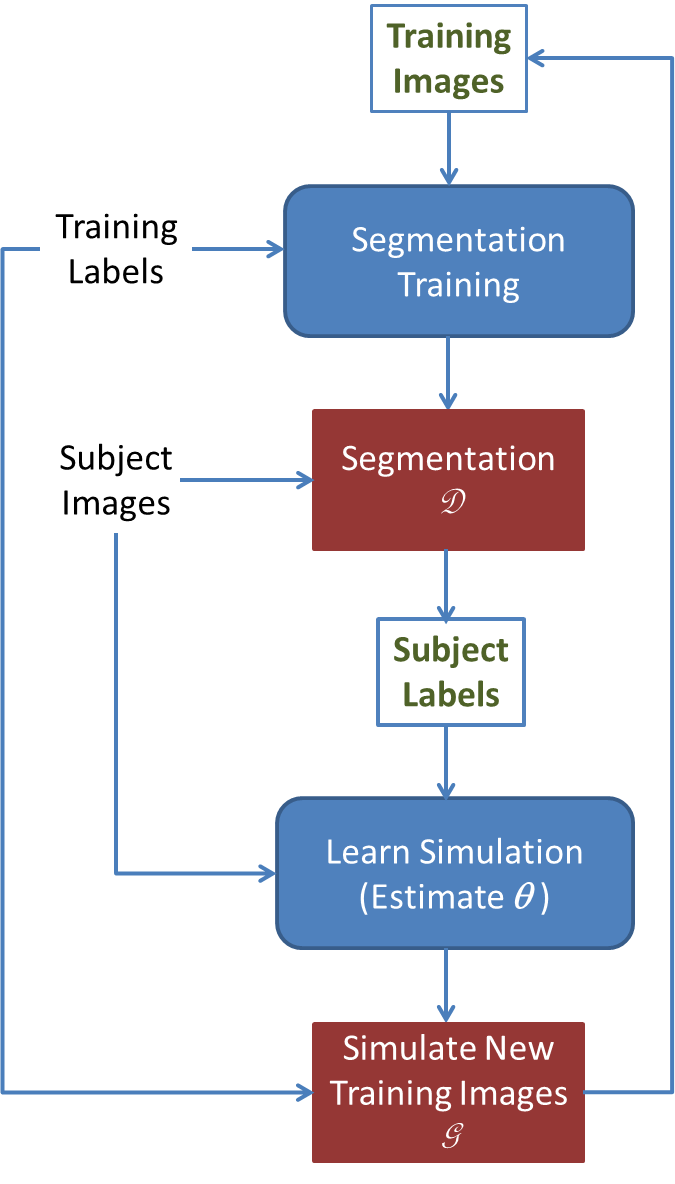}
\caption{\footnotesize Block diagram of contrast adaptive framework}
\label{fig:block}
\end{center}
\vspace{-0.5in}
\end{wrapfigure} 

In this work, we propose a novel framework to directly integrate a generative MR simulation into the
segmentation process.  We use the MR simulation process to generate new images within the training
data from the label images.  As a proof of concept, we adopt a simple simulation model and show that
by minimizing the error between the input image and its simulated version based on the current
segmentation estimate, contrast differences between the training and input images can be effectively
addressed.   Preliminary results applying the approach to synthetic and real MR images demonstrate
substantial improvement in obtaining a consistent segmentation, on the same level as when the training and
input images have the same contrast.  Our approach bares conceptual similarites to generative adversarial
networks~[\citenum{goodfellow14}], which also employs both generative and discriminative processes,
but the rationale and implementation are different here.

\section{Methods}

We denote a segmentation $\mathcal{D}(\cdot)$ to be a mapping from the MR image to its anatomical
labels, which represent the desired regions of interest.  We consider the segmentation to be a
supervised process that requires a training data set consisting of MR image and label image pairs.
We denote the simulation $\mathcal{G}(\cdot;\bm{\theta})$ to be a mapping from the labels to an MR
image, parameterized by the vector $\bm{\theta}$.  Our combined segmentation and simulation
framework is illustrated in Fig.~\ref{fig:block}.  Note that the training label images and subject
images are fixed, while the training MR images and subject labels are updated with each iteration.
In the first iteration, the segmentation is performed as usual with the available training data,
yielding an initial segmentation estimate.  This estimate is then used to compute the parameters
within the simulation process by minimizing a consistency equation.  Given the input MR image $X$
and segmentation estimate $L$, we estimate $\bm{\theta}$ according to
\begin{equation}
\hat{\bm{\theta}} = \arg \min_{\theta} \| X - \mathcal{G}(L;\bm{\theta})\|^2 
\label{eq:min}
\end{equation}
Using the estimated simulation parameters, we generate new MR images from the label images within
the training data, which are subsequently used to segment the input image again.  This process
iterates until minimal changes occur within the segmentation.

Because the purpose of this work is to primarily serve as a proof of concept of this approach, we
use a relatively simple classifier for $\mathcal{D}$ and a simple simulation model for
$\mathcal{G}$.  Although more sophisticated models, such as deep learning networks, can be used, our
simplified choices allow the proposed framework to be more conveniently demonstrated.  For
$\mathcal{D}$, we use a Gaussan classifier for segmentation.  A Gaussian classifer in this context
models each tissue class as a Gausian distribution, given by,
\begin{equation}
p_{jk} = \frac{1}{\sqrt{2\pi\sigma^2}}\exp{\frac{(x_j-\mu_k)^2}{2\sigma_k^2}}
\label{eq:gauss}
\end{equation}
where $\mu_k$ and $\sigma_k^2$ are the mean and variance of the intensities for each tissue class
$k$ and are estimated from the training data.  The variable $j$ represents the pixel or voxel
location.  A standard segmentation approach would use the estimated values of $\mu_k$ and
$\sigma_k^2$ from the training data to compute the probability that each voxel in the input image
$X$ belongs to tissue class $k$ and stop at that point.  A hard label image can be generated by
assigning each voxel to the tissue class of highest probability.

Under our framework, we alternate the classification step with a simulation step.  Given the input
image, and an estimate of the segmentation probability functions, we seek to define the simulation
process that maps the segmentation to the intensities.  For the simulation model, we assume that
each pixel is a linear combination of a tissue class centroid weighted by the probability or
partial volume fraction.  This is given by
\begin{equation}
y_j = \sum_{k=1}^K p_{jk}c_k
\label{eq:sim}
\end{equation}
where $y_j$ is the pixel or voxel intensity at location $j$, $p_{jk}$ is the probability or partial
volume fraction that voxel $j$ belongs to label or class $k$, and $c_k$ is the intensity of label
$k$.  For simplicity, noise effects are ignored in the model.  

Under this model, $\bm{\theta} =
[c_1,\ldots,c_k]^T$ and it is straightforward to show that Eq.~\ref{eq:min} is solved by
\begin{equation}
\bm{\theta} = (P^TP)^{-1}P^T\bm{x}
\end{equation}
where $P$ is the matrix of probabilities $p_{jk}$, and $\bm{x}$ is a stacking of the input MR image
pixels $x_j$.  For estimating $\bm{\theta}$, the probabilities are taken from the Gaussian
classifier segmentation result.  We note that $c_k$ is computed from the input image, while $\mu_k$
is computed from the training data.  For generating new MR images in the training data, we apply
Eq.~\ref{eq:sim} assuming that the probabilities are provided within the label images of the
training data, or, if unavailable, derived from a blurring filter applied to the hard labels.

\subsection{New or breakthrough work to be presented}
The inconsistent performance of supervised segmentation algorithms when applied to images with
different contrast properties is a critical issue in medical image analysis.  This problem manifests
itself in two common ways.  First, if the contrast of an input image is different from that of the
training data, performance is compromised, generally leading to reduced segmentation accuracy and
rendering tools for detecting subtle differences in anatomical structure unusable.  Second, when
attempting to analyze data acquired with different protocols or from from different sites, standard
approaches will lead to significant biases in the segmentation results that prohibit direct comparisons
and pooling of data.  This work helps restore the ability to find subtle relationships in
pooled heterogeneous MR data sets by combining segmentation and simulation techniques.  Although previous
methods have been proposed for contrast matching, our  approach benefits from being
integrated into the segmentation process.  This work has not been presented elsewhere.

\section{Results}
\label{sec:results}

\begin{figure}[t]
\centering
\begin{minipage}{0.95\textwidth}
\tabcolsep 2pt
\begin{tabular}[ht]{cccccc}
\includegraphics[height=2.6cm]{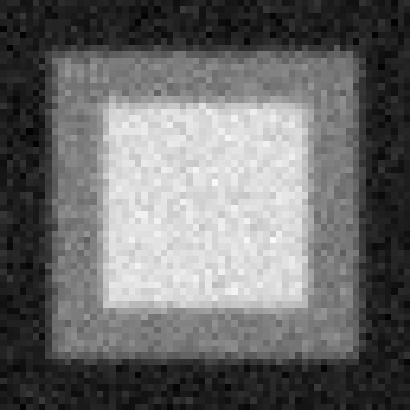}& 
\includegraphics[height=2.6cm]{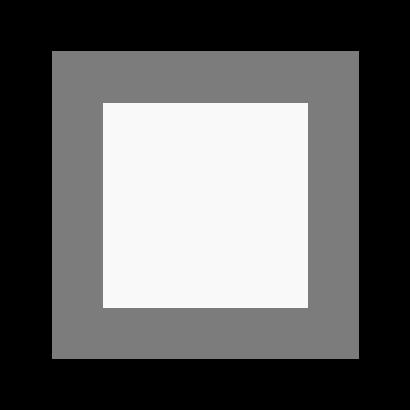}&
\includegraphics[height=2.6cm]{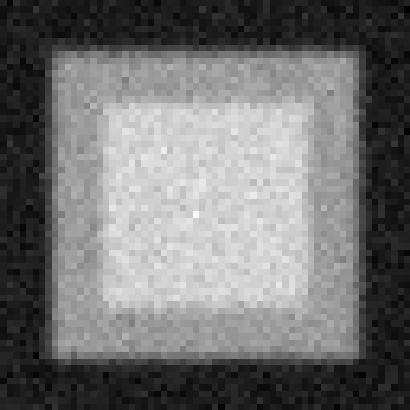}&
\includegraphics[height=2.6cm]{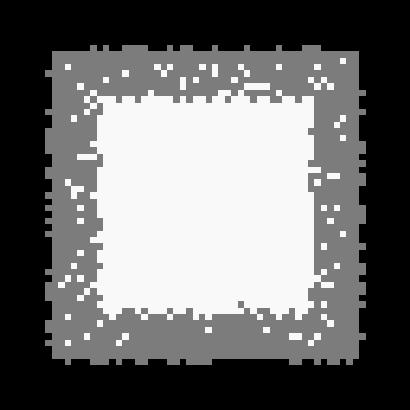}&
\includegraphics[height=2.6cm]{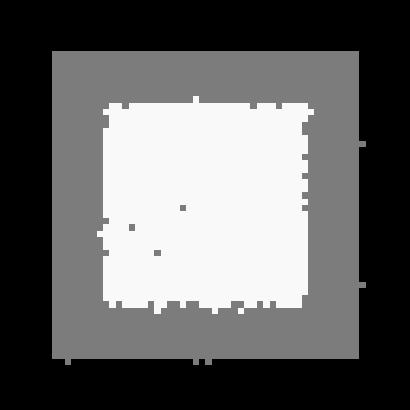}&
\includegraphics[height=2.6cm]{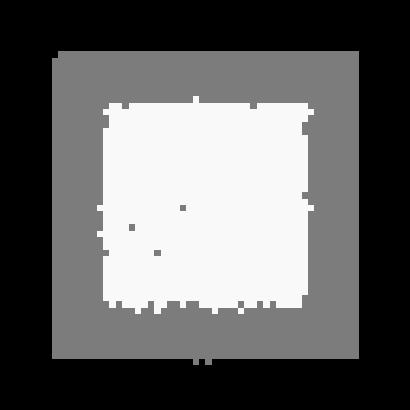} \\
(a) & (b) & (c) & (d) & (e) & (f)\\
%\tiny Training Image & \tiny Training Labels &
%\tiny Test Image & \tiny Standard Segmentation & 
%\tiny Proposed Segmentation &  \tiny Ideal Segmentation\\
\end{tabular}
\end{minipage}
\caption{Segmentation results on a synthetic test image with 3 classes: (a) training image, (b)
  training labels, (c) test image, (d) standard segmentation, (e) proposed segmentation, (f) ideal segmentation.}
%\vspace{-0.1in}
\label{fig:synthetic_image}
\end{figure}

\begin{figure}
%\begin{wrapfigure}{r}{0.4\textwidth}
%\vspace{-0.2in}
\begin{center}
\includegraphics[width=11cm]{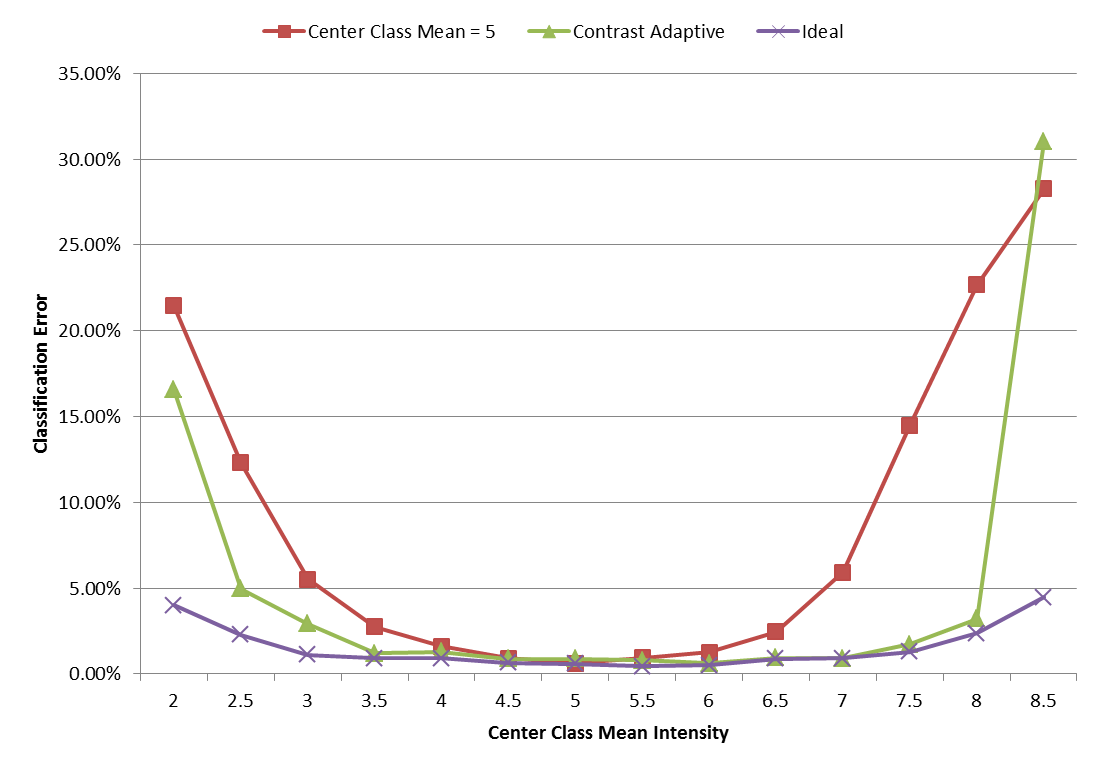}
\caption{Classification errors as the mean of the center class from Fig.~\ref{fig:synthetic_image}
  varies from 2 to 8.5.  The purple curve represents performance with ideal training data that
  varies according to the input.  The green curve represents our proposed approach, and the red
  curve represents performance when the training data is fixed. }
\label{fig:error}
\end{center}
\vspace{-0.1in}
%\end{wrapfigure} 
\end{figure}

We first demonstrate our method using a synthetic test image shown in
Fig.~\ref{fig:synthetic_image}.  The training image consisted of 3 classes with means 0, 5, and 10,
and included blurring and additive Gaussian noise with a standard deviation of 0.5.  The test image
had the center class mean altered, and in the case of Fig.~\ref{fig:synthetic_image}(c), set to 7
rather than 5.  Applying a Gaussian classifer to the test image yielded poor results, as shown
Fig.~\ref{fig:synthetic_image}(d).  Using our contrast adaptive framework, five iterations were
required to convergence (defined as less 0.01\% pixels changing labels).  The resulting segmentation
in Fig.~\ref{fig:synthetic_image}(e) is very similar to the result one would obtain if the
classifier were trained on data with the same contrast properties as the input, which we refer to as
the ``ideal'' segmentation shown in Fig.~\ref{fig:synthetic_image}(f).

Figure~\ref{fig:error} shows how the classification error, defined as the ratio of misclassified
pixels to total pixels, varies as the center class mean is changed from 2 to 8.5.  The bottom curve
represents the ideal segmentation error, where the training data has the same contrast as the test
data.  The top curve represent the scenario where the training data is fixed, with the center class
mean equal to 5.  The curve in between represents our proposed algorithm, which reduces the
classification error as the class mean moves away from 5.  Note also that the performance of our
algorithm is better at higher intensity values than lower ones.  This is because at lower mean
intensities, the contrast to noise ratio deteriorates more rapidly, with the magnitude of noise
becoming almost equal to the magnitude of the signal difference between each class.  Nevertheless,
there does reach a point where the simulation step does not improve segmentation results and the
error increases dramatically.

\begin{figure}[t]
\centering
\begin{minipage}{0.95\textwidth}
\tabcolsep 2pt
\begin{tabular}[ht]{cccccc}
\includegraphics[height=2.6cm]{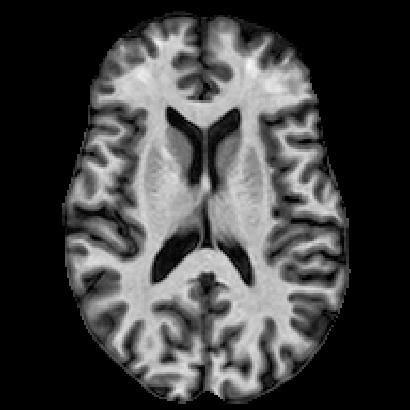}& 
\includegraphics[height=2.6cm]{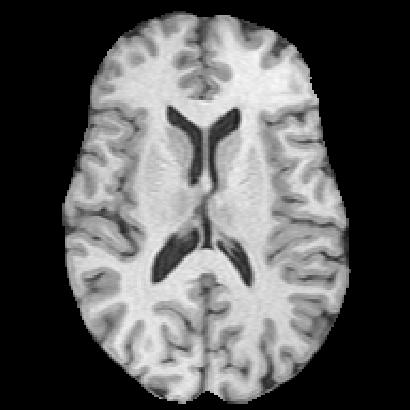}&
\includegraphics[height=2.6cm]{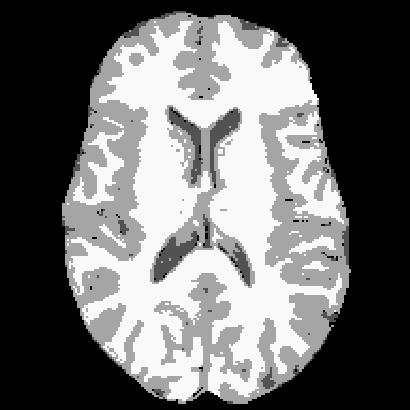}&
\includegraphics[height=2.6cm]{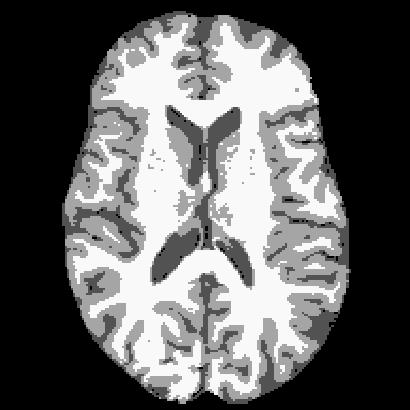}&
\includegraphics[height=2.6cm]{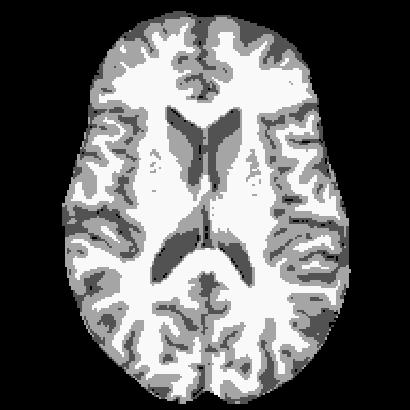}&
\includegraphics[height=2.6cm]{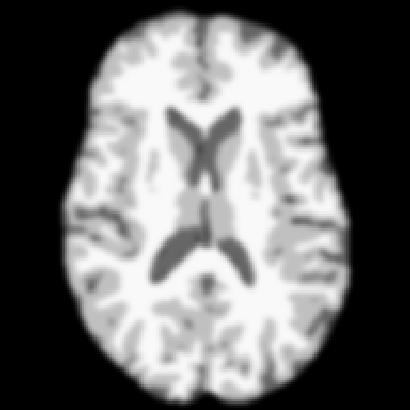} \\
(a) & (b) & (c) & (d) & (e) & (f)\\
%\tiny Training Image & \tiny Training Labels &
%\tiny Test Image & \tiny Standard Segmentation & 
%\tiny Proposed Segmentation &  \tiny Ideal Segmentation\\
\end{tabular}
\end{minipage}
\caption{Segmentation results on MR images with different contrast properties: (a) MPRAGE training
  image, (b) SPGR test image, (c) standard segmentation of SPGR, (d) proposed segmentation of SPGR, (e) segmentation
of MPRAGE image, (f) simulated SPGR image.}
\vspace{-0.1in}
\label{fig:mri}
\end{figure}

Fig.~\ref{fig:mri} shows the effect of applying our proposed approach to segment a T1-weighted MR
image.  Figs.~\ref{fig:mri}(a) and (b) show co-registered T1-weighted images acquired from the same
subject.  The former image is an MPRAGE acquisition, and the latter image is an SPGR, each leading
to a rather distinct depiction of the brain anatomy.  The images have been pre-processed for
inhomogeneity correction and brain extraction.  Using the MPRAGE as a training image for a
three-class, supervised Gaussian classifier applied to the SPGR yields the result shown in
Fig.~\ref{fig:mri}(c).  The lateral ventricles appear to be under segmented and sulcal cerebrospinal
fluid (CSF) is almost entirely lacking.  Using the proposed algorithm, one achieves a very similar
segmentation result to what would be obtained if directly segmenting the MPRAGE image, as shown in
Figs.~\ref{fig:mri}(d) and (e).  The Dice coefficents between these two segmentations are 0.81,
0.81, and 0.91 for CSF, gray matter, and white matter respectively.  Comparing the MPRAGE
segmentation to the standard SPGR segmentation, the Dice coefficents are 0.42, 0.69, and 0.88.  This
amounts to an overall improvement of over 27\% in segmentation consistency.  The consistency in
total brain (gray and white matter) estimated from the segmentations of the SPGR and the MPRAGE
images without and with contrast adaptation is improved from 17.5\% to 0.7\%, the latter value being
within the range of typical segmentation error~\cite{Heinen16}.

Fig.~\ref{fig:mri}(f) shows the final simulated MR image within the training data.  The overall
contrast of the training data has been transformed to be more similar to the SPGR, although there is
ample room for improving the realism of the simulation.  In particular, the
simulation model employed here is quite rudimentary in comparison to other techniques that have been
proposed, such as in~[\citenum{He15}].

\subsection*{Conclusions}
Our preliminary data demonstrates improved robustness to contrast differences between training and
test images when using our proposed framework.  Better performance can likely be achieved by using
more sophisticated techniques for both the segmentation and simulation.  In future work, we will use
deep learning networks for both the segmentation and simulation steps.  A limitation of this work is
that a whole brain segmentation in order to produce reasonable simulations.  Its use with
lesion detection techniques, for example, would require significant modifications.  

\subsection*{Acknowledgments}
This work was supported by the Department of Defense in the Center for Neuroscience and Regenerative
Medicine, and by grant RG-1507-05243 from the National Multiple Sclerosis Society.  

%   \begin{figure} [ht]
%   \begin{center}
%  \begin{tabular}{c} 
% \includegraphics[height=5cm]{MultimediaFigure.jpg}
%	\end{tabular}
%	\end{center}
 %  \caption[example] 
 % { \label{fig:video-example} 
%A label of “Video/Audio 1, 2, …” should appear at the beginning of the caption to indicate to which multimedia file it is linked . Include this text at the end of the caption: \url{http://dx.doi.org/doi.number.goes.here}}
 %  \end{figure} 

% References
\bibliography{contrast} % bibliography data in report.bib
\bibliographystyle{spiebib} % makes bibtex use spiebib.bst

\end{document}